%% file: neurips_2024.tex
\documentclass{article}

\usepackage[preprint]{neurips_2025}

\usepackage[utf8]{inputenc} 
\usepackage[T1]{fontenc}    
\usepackage{hyperref}       
\usepackage{url}            
\usepackage{booktabs}       
\usepackage{amsfonts}       
\usepackage{nicefrac}       
\usepackage{microtype}      
\usepackage{xcolor}         
\usepackage{graphicx}  
\usepackage{adjustbox} 
\usepackage{pifont}    
\usepackage{tabularx}  
\usepackage{makecell}  
\usepackage{algorithm}
\usepackage{algorithmic}
\usepackage{amsmath}
\usepackage{booktabs}
\usepackage{longtable}
\usepackage{booktabs}
\usepackage{multirow}
\usepackage{adjustbox}
\usepackage{subcaption}
\usepackage{floatflt}
\usepackage{xcolor}
\usepackage{listings}
\usepackage{makecell} 

\title{Bidirectional LMs are Better Knowledge Memorizers? A Benchmark for Real-world Knowledge Injection}

\author{%
  Yuwei Zhang$^1$ \quad Wenhao Yu$^2$ \quad Shangbin Feng$^3$ \quad Yifan Zhu$^1$ \quad Letian Peng$^1$ \\
  \bf Jayanth Srinivasa$^4$ \quad Gaowen Liu$^4$ \quad Jingbo Shang$^1$ \\
  UC, San Diego$^1$ \quad Tencent AI Lab Seattle$^2$ \quad
  University of Washington$^3$ \quad Cisco$^4$
}

\begin{document}

\maketitle

\newcommand{\dataset}{\textsc{WikiDYK}}

\begin{abstract}
    Despite significant advances in large language models (LLMs), their knowledge memorization capabilities remain underexplored, due to the lack of standardized and high-quality test ground.
    In this paper, we introduce a novel, real-world and large-scale knowledge injection benchmark that evolves continuously over time without requiring human intervention.
    Specifically, we propose {\dataset}, which leverages recently-added and human-written facts from Wikipedia's ``Did You Know...'' entries. These entries are carefully selected by expert Wikipedia editors based on criteria such as verifiability and clarity.
    Each entry is converted into multiple question–answer pairs spanning diverse task formats from easy cloze prompts to complex multi-hop questions.
    {\dataset} contains $12,290$ facts and $77,180$ questions, which is also seamlessly extensible with future updates from Wikipedia editors.
    Extensive experiments using continued pre-training reveal a surprising insight:
    despite their prevalence in modern LLMs, Causal Language Models (CLMs) demonstrate significantly weaker knowledge memorization capabilities compared to Bidirectional Language Models (BiLMs), exhibiting a $23\%$ lower accuracy in terms of reliability.
    To compensate for the smaller scales of current BiLMs, we introduce a modular collaborative framework utilizing ensembles of BiLMs as external knowledge repositories to integrate with LLMs. Experiment shows that our framework further improves the reliability accuracy by up to $29.1\%$.
    
\end{abstract}

\input{Sections/1_Intro}

\input{Sections/2_Related}
\input{Sections/3_Data}

\input{Sections/4_Train}

\input{Sections/5_Exp}

\input{Sections/6_Discuss}

\bibliographystyle{apalike}
\bibliography{ref}

\appendix

\input{Sections/7_appendix}

\end{document}

%% file: Sections/1_Intro.tex
\section{Introduction}

Large language models (LLMs) acquire most of their knowledge during pre-training, by learning patterns from massive web-scale corpora~\citep{chang2024large,chen2024continual,li2025memorization}. This process allows them to recall facts, reason over information, and generate coherent text without explicit supervision. As a result, LLMs are often treated as static knowledge bases -- capable of answering factual queries based on what they have seen during training. However, this raises an important question: \textit{Can LLMs truly memorize and internalize new knowledge after pretraining?}
Previous work suggests significant challenges in effectively updating the internal knowledge of language models~\citep{jang2021towards,jang2022temporalwiki,kim2023carpe,ovadia2023fine,fu2023revisiting,zhang2023plug,mecklenburg2024injecting,xu2025memorizing}. However, these findings are largely based on synthetic datasets derived from noisy Wikipedia snapshots where the knowledge may lack real-world significance and inherent complexities. Moreover, evaluations using synthetic questions often suffer from ill-defined contexts, such as ambiguous queries like "What is the value of y?", undermining their effectiveness in accurately reflecting model performance.

\newcommand{\cmark}{\ding{51}} 
\newcommand{\xmark}{\ding{55}} 
\begin{table}[t]
    \centering
    \caption{Comparison of {\dataset} with existing benchmarks for knowledge injection.}
    \label{tab:wikidyk_comparison_transposed}
    \begin{adjustbox}{max width=\textwidth}
    \setlength{\tabcolsep}{2mm}{
    \begin{tabular}{lcccccc}
        \toprule
        \textbf{Benchmark} 
            & \makecell{\textbf{Dataset} \\ \textbf{Source}} 
            & \makecell{\textbf{Topical} \\ \textbf{Scope}} 
            & \makecell{\textbf{Update} \\ \textbf{Frequency}} 
            & \makecell{\textbf{Human} \\ \textbf{Curated}} 
            & \makecell{\textbf{Automatic} \\ \textbf{Extension}} 
            & \makecell{\textbf{Available} \\ \textbf{Tasks}} \\
        \midrule
        EvolvingQA \citep{kim2023carpe} 
            & Wikipedia     & General  & \textasciitilde80K monthly 
            & \xmark        & \cmark        & QA            \\
        RealtimeQA \citep{kasai2023realtime} 
            & News    & News & 30 weekly   & \cmark  & \xmark  & QA      \\
        StreamingQA \citep{liska2022streamingqa} 
            & News    & News & 9K quarterly   & hybrid  & \xmark  & QA       \\
        TemporalWiki \citep{jang2022temporalwiki} 
            & Wikipedia    & General & 300K monthly & \xmark       & \cmark       & Slot-filling \\
        CKL \citep{jang2021towards} 
            & Pre-train Data & General & – & hybrid & \xmark & Slot-filling \\
        {\dataset} (Ours) 
            & Wikipedia & General & \textasciitilde10 daily  & \cmark           & \cmark           & QA               \\
        \bottomrule
    \end{tabular}}
    \end{adjustbox}
    \vspace{-3mm}
\end{table}

\begin{figure}[t]
\setlength{\floatsep}{6pt plus 2pt minus 2pt}
    \centering
    \includegraphics[width=\linewidth]{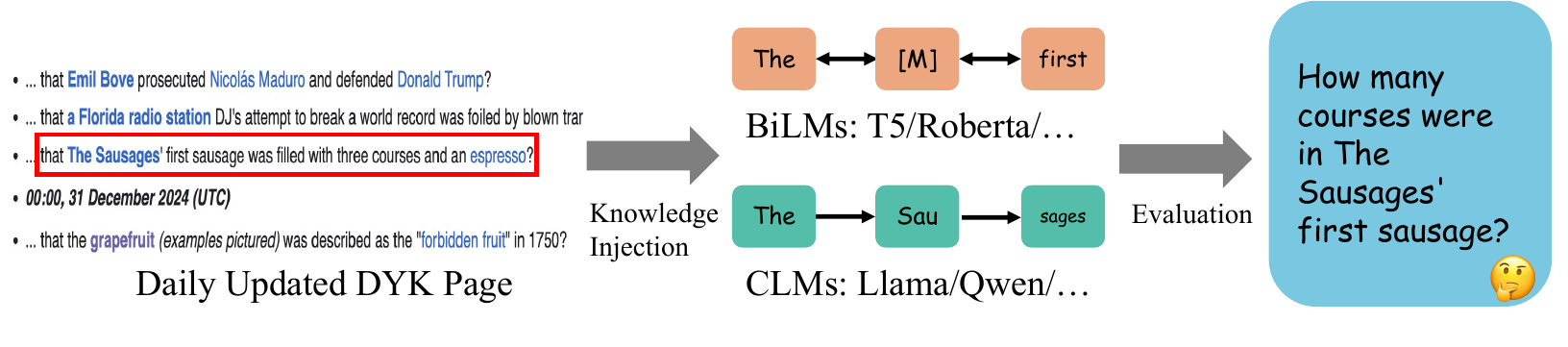}
    \vspace{-6mm}
    \caption{Proposed knowledge injection evaluation workflow. We inject the knowledge from via continued pre-training which can be achieved via various model architectures or training objectives. The injected model is then evaluated with questions from multiple dimensions from easy cloze prompts to complex multi-hop questions. Notice that the images are not used in the dataset.}
    \label{fig:workflow}
    \vspace{-6mm}
\end{figure}

We extend previous efforts on knowledge injection by building a novel large-scale and high-quality benchmark -- {\dataset} -- derived from a natural, expert-curated, and constantly updating knowledge source: Wikipedia’s “Did You Know...” (DYK) pages\footnote{\url{https://en.wikipedia.org/wiki/Wikipedia:Did_you_know}}. These pages highlight Wikipedia's continuous growth and domain diversity by featuring daily updates of facts reviewed by expert Wikipedia editors. Each day, about $10$ facts are added to the list, selected from recently expanded articles which likely not exist in pre-training data while adhering to Wikipedia's most important content policies (\emph{e.g.} ``Mountain lions in the Santa Monica Mountains of Los Angeles are one of only two examples of wild big cats living in a megacity''). {\dataset} leverages this structured, human-driven process to ensure both novelty and quality, offering a unique resource for evaluating knowledge injection in language models that go beyond synthetic dataset construction. We compare our {\dataset} with previous benchmarks in Table~\ref{tab:wikidyk_comparison_transposed}.

With {\dataset} we aim to systematically evaluate the performance of LLM knowledge injection. Building on prior work in knowledge editing~\citep{meng2022locating,meng2022mass, wang2023easyedit}, we design a multi-dimensional evaluation suite using open-domain QA, spanning lower-level knowledge memorization to higher-level knowledge association tasks~\citep{xu2025memorizing}. To guarantee the ease of extension in the future, all evaluation questions are generated via a lightweight prompt-based method that employs only the factual knowledge and corresponding Wikipedia articles. An example of our evaluation workflow is visualized in Figure~\ref{fig:workflow}.
\begin{floatingfigure}[r]{0.5\textwidth}
  \centering
  \includegraphics[width=0.49\textwidth]{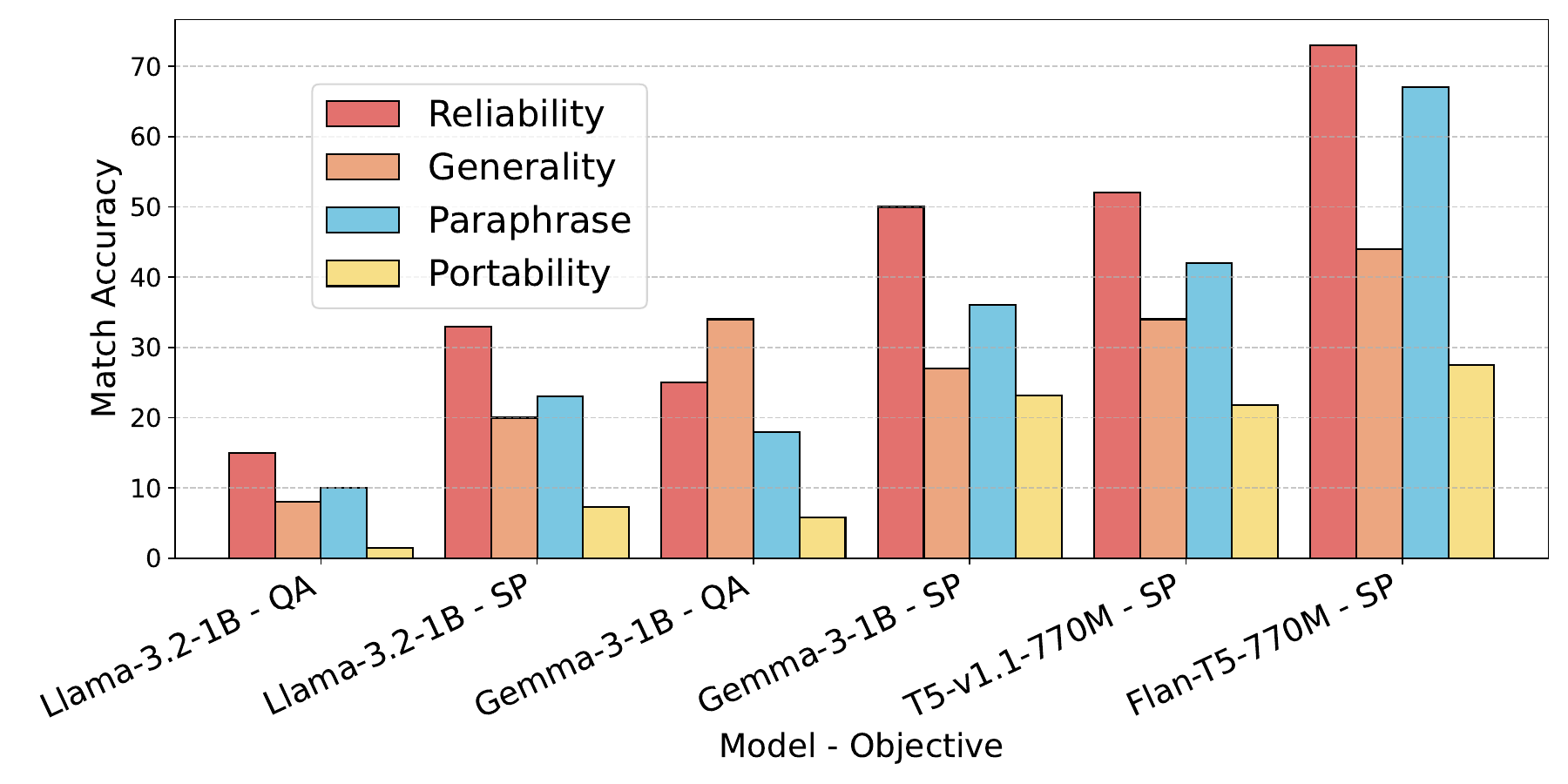}
  \vspace{-1mm}
  \caption{Evaluation results of injecting $1,000$ facts from {\dataset}. We report performance for two CLMs and two BiLMs. A more comprehensive set of results can be found in Appendix~\ref{sec:more_results}.}
  \label{fig:results_barplot}
  \vspace{-1mm}
\end{floatingfigure}
Based on these designs, we conduct a comprehensive comparative analysis of both model architectures and training objectives for knowledge injection. 
Given the observations on the ``reverse curse'' issue of Causal Language Models' (CLMs)~\citep{berglund2023reversal}, we aim to investigate whether Bidirectional Language Models (BiLMs) that leverage the context from both sides can generalize better.
Notably, our experiments reveal that the smaller BiLMs significantly outperforms the most recent large CLMs at memorizing knowledge as shown in Figure~\ref{fig:results_barplot}, even after aligning the training objective and the amount of training.
We attribute this discrepancy to the reduced context visibility in CLMs during training, which hinders their ability to efficiently encode factual knowledge.
In fact, bidirectional attention has also been found to facilitate model editing and finetuning~\citep{ma2023untying,kopiczko2024bitune}.
Finally, to address the challenges of scaling knowledge injection while mitigating catastrophic forgetting, we propose a framework that leverages BiLMs as dedicated knowledge repositories.
These are dynamically integrated with LLMs through a scope classifier acting as an adaptive router. We showcase on {\dataset} that it is possible combine the vast pretrained knowledge with newly injected of BiLMs effectively. Our framework can also be utilized on applications that require integrating knowledge from multiple domains and enables efficient updates via localized retraining, avoiding full-model retraining overhead.

In summary, our contribution is two-fold: a high-quality and expert-curated dataset {\dataset} to evaluate LLM knowledge injection, and a call-to-action to revisit bidirectional LMs for neural knowledge modeling.
Code and data are publicly available: \url{https://github.com/zhang-yu-wei/WikiDYK}; \url{https://huggingface.co/datasets/YWZBrandon/wikidyk}.

%% file: Sections/2_Related.tex
\section{Related Works}

\noindent\textbf{Wikipedia's Did You Know (WikiDYK).} WikiDYK is a section of Wikipedia that features newly added or expanded facts, updated daily to showcase novel and verified information. The XQA dataset~\citep{liu2019xqa} uses the multilingual feature of WikiDYK sections to automatically generate multilingual question-answer pairs by masking named entities in factual statements. \cite{rybak2020klej} takes the Polish split of WikiDYK to benchmark Polish question answering systems. \cite{prakash2015did} treat WikiDYK as trivia for entities inside Wikipedia to structure the relations between Wikipedia entities. Unlike previous work, we explicitly leverage the temporal structure of WikiDYK to analyze how well language models can incorporate and update newly emerging knowledge over time, enabling dynamic evaluation of knowledge injection effectiveness across different temporal slices.

\noindent\textbf{{Temporal Evolution of LLMs.}} Large language models (LLMs) have a fixed temporal knowledge scope, known as the knowledge cutoff~\citep{cheng2024dated}, making knowledge injection crucial for maintaining relevance~\citep{song2025injecting}. Supervised fine-tuning (SFT) has proven effective, especially when using fact-based rather than token-based data to incorporate recent, domain-specific information~\citep{mecklenburg2024injecting}, though retention remains a challenge~\citep{ovadia2023fine}. Surprisingly, injecting unaligned or random knowledge can perform as well as aligned data, prompting calls for better pruning and refinement methods~\citep{fu2023revisiting}. To improve efficiency, plug-and-play approaches allow external knowledge to be injected into frozen models via map-tuned embeddings~\citep{zhang2023plug}. While retrieval-augmented generation (RAG) still outperforms fine-tuning in handling novel facts~\citep{ovadia2023fine}, future work should integrate retrieval and fine-tuning with improved alignment and parameter efficiency for robust, up-to-date reasoning.

\noindent\textbf{{Generalization of Injected Knowledge.}} Once knowledge is injected, LLMs are expected not only to memorize the fine-tuned sequences but also to generalize the new information across diverse contexts. However, numerous studies in knowledge editing have underscored a central challenge: standard fine-tuning methods often struggle to simultaneously meet multiple critical objectives~\citep{meng2022locating, onoe2023can, hoelscher2023detecting, gupta2023editing}. To better understand how injected knowledge influences generalization, researchers have explored fine-tuning learning dynamics~\citep{LearningDynamics}, applied influence functions~\citep{llm_influence_function}, and analyzed representation correlations~\citep{lm_correlation}. While a range of benchmarks has been developed to assess generalization in this setting~\citep{kim2023carpe, kasai2023realtime, liska2022streamingqa, jang2022temporalwiki, jang2021towards}, many fall short in terms of quality and complexity compared to WikiDYK.

%% file: Sections/3_Data.tex
\begin{table}[ht!]
\centering
\footnotesize
\setlength{\tabcolsep}{4pt}
\renewcommand{\arraystretch}{0.98}
\caption{Illustrative example questions drawn from our multi-dimensional evaluations. Please refer to Section~\ref{sec:evaluation} for an in-depth explanation of each dimension and task.}
\label{tab:example}
\begin{adjustbox}{max width=0.95\textwidth}
\begin{tabularx}{\linewidth}{@{}lX@{}}
\toprule
\textbf{Type}        & \textbf{Example} \\
\midrule
Original        & Kanye West originally wrote the chorus of “\textbf{Gold Digger}” from a female point of view. \\
\addlinespace
Reliability          & What is the name of the song for which Kanye West originally wrote the chorus from a female point of view? Answer: Gold Digger \\
\addlinespace
Generality           & From whose point of view did Kanye West originally write the chorus of ``Gold Digger''? Answer: female \\
\addlinespace
Paraphrase           & What is the title of the song where Kanye West initially penned the chorus from a woman's perspective? Answer: Gold Digger \\
\addlinespace
Portability          & I recently came across a story about an American artist, born in 1977, who reshaped hip-hop with his ever-evolving sound and innovative style. I heard he once wrote a chorus meant to be sung from a female perspective for one of his tracks. Do you know which song that was? Answer: Gold Digger \\
\addlinespace
Locality             & Which American artist, born in 1977, revolutionized hip-hop with innovative music and influential fashion ventures, and is known for both his Grammy-winning albums and controversial public persona? Answer: Kanye West \\
\bottomrule
\end{tabularx}
\end{adjustbox}
\vspace{-5mm}
\end{table}

\section{{\dataset} Benchmark}\label{sec:data}

We introduce the dataset collection and analysis in this section. Our dataset is constructed based on the daily updated Wikipedia's recent additions website \footnote{\url{https://en.wikipedia.org/wiki/Wikipedia:Recent_additions}} which contains numerous expert-reviewed new knowledge.
In order to facilitate timely updates, we design a simple prompt-based QA generation workflow that can be used to evaluate knowledge injection models from various difficulty levels.

\subsection{Data Collection}

On the ``DYK'' webpage, about $10$ facts are selected each day from either recently created new articles or greatly expanded existing articles. The bolded entity (\emph{e.g.} ``Gold Digger'') within the fact is linked to the original article that introduces the new knowledge. We scrape both the raw text and the accompanying Wikipedia article with additional cleaning from the webpage. In order to ensure the knowledge is up-to-date and align with LLM knowledge cutoffs, we only acquire those pages starting from January 2022 until April 2025 with a total number of $12,290$ facts.

\subsection{Evaluation}\label{sec:evaluation}

\noindent \textbf{Question Generation.} We construct question-answer (QA) pairs using two sources: scraped factual knowledge snippets and Wikipedia articles. To comprehensively evaluate knowledge memorization and association capabilities~\citep{xu2025memorizing}, we design multi-dimension evaluation including five distinct question types as detailed in Table~\ref{tab:example}.
\textbf{Reliability}: Directly tests recall of bolded entities by formulating questions from the non-bolded context of the original fact (\emph{e.g.}, ``What song...'' for the bolded Gold Digger).
\textbf{Generality}: Extracts answers from implicit non-bolded components of the same fact (\emph{e.g.}, inferring female from ``female point of view'') which tests whether the injected model can accurately recall knowledge.
\textbf{Paraphrase}: Uses syntactically rephrased or lexically substituted versions of the original fact (e.g., from ``initially penned" to "originally wrote").
\textbf{Locality}: Evaluates retention of pre-trained knowledge (\emph{e.g.}, biographical details) after new knowledge injected, ensuring no catastrophic forgetting. Specifically, a question is generated based on the description of an entity other than the bolded one in the fact (\emph{e.g.} ``Kanye West'' in Table ~\ref{tab:example}).
\textbf{Portability}: Requires multi-hop reasoning between injected knowledge (e.g., "chorus written from a female perspective") and pretrained knowledge (e.g., "groundbreaking artist... experimental beats"). See Appendix~\ref{sec:prompts} for question generation prompts.

\noindent \textbf{Metrics and Models.} We use substring match accuracy and token F1 as our evaluation metrics following the convention in open-domain QA. A simple question template is applied to facilitate answer generation: ``\{question\}\textbackslash nAnswer:''. For evaluation, we choose $6$ open-source LLMs from $3$ different model families and scales. See more detailed description in Section~\ref{sec:exp_setup} and Appendix~\ref{sec:hyper}.

\subsection{Static Analysis}\label{sec:data_analysis}

We first evaluate the zero-shot performance of off-the-shelf language models on {\dataset} prior to knowledge injection (termed static performance), with results in Table~\ref{tab:merged_static_rag}. Despite these models reporting knowledge cutoffs extending to 2023, they exhibit near-chance accuracy on {\dataset}, indicating the novelty of the provided knowledge. In contrast, performance on locality questions—which probe pretraining knowledge—is significantly higher, aligning with expectations for static model behavior. For completeness, we include static results on reliability questions from 2004\textendash{}2009 in Appendix~\ref{sec:more_results}.

\subsection{Impact of Retrieval Augmented Generation (RAG)}
We further analyze the impact of RAG-augmented models in Table~\ref{tab:merged_static_rag} where we use all the collected Wikipedia articles as our retrieval data store and use a popular sentence embedding model\footnote{\url{BAAI/bge-small-en-v1.5}} to retrieve top-$k$ articles. While RAG consistently improves performance, its practical application in knowledge injection faces challenges: (1) the computational overhead of retrieving and processing external contexts remains prohibitive for applications that are latency-sensitive, and (2) reliance on external datastores complicates deployment pipelines and introduces potential privacy risks. These limitations underscore the importance of effective knowledge injection methods.

\begin{table}[t]
\centering
\caption{Performance comparison between static models and \texttt{+RAG}. RAG retrieves top-3 Wikipedia articles per question. \texttt{Llama-2-7b} and \texttt{Flan-T5} are excluded from RAG due to context length limits.}
\label{tab:merged_static_rag}
\begin{adjustbox}{max width=0.9\textwidth}
\begin{tabular}{l *{5}{cc}}
\toprule
\multirow{2}{*}{\textbf{Model}}
  & \multicolumn{2}{c}{\textbf{Reliability}}
  & \multicolumn{2}{c}{\textbf{Generality}}
  & \multicolumn{2}{c}{\textbf{Paraphrase}}
  & \multicolumn{2}{c}{\textbf{Portability}}
  & \multicolumn{2}{c}{\textbf{Locality}} \\
\cmidrule(lr){2-3} \cmidrule(lr){4-5}
\cmidrule(lr){6-7} \cmidrule(lr){8-9} \cmidrule(lr){10-11}
  & Match & F1 & Match & F1 & Match & F1 & Match & F1 & Match & F1 \\
\midrule
Flan-T5-220M      &  0.15 & 3.27 & 2.50 & 5.88 & 0.11 & 3.33 & 0.19 & 2.84 & 5.46 & 14.22 \\
Flan-T5-770M     &  0.23 & 3.70 & 3.39 & 7.68 & 0.24 & 3.80 & 0.25 & 3.15 & 10.47 & 16.45 \\
Llama-2-7b        &  1.30 & 0.99 &  5.84 & 1.12 &  1.34 & 0.97 &  0.86 & 0.56 & 51.36 &  5.56 \\
\addlinespace[2pt]
Llama-3.1-8B      &  1.94 & 1.05 &  7.68 & 1.30 &  2.18 & 1.06 &  1.00 & 0.59 & 58.52 & 11.51 \\
\multicolumn{1}{r}{+RAG}  & 25.85 &12.50 & 30.32 &11.82 & 25.81 &12.62 & 20.98 & 5.54 & 40.86 & 16.19 \\
\addlinespace[2pt]
Llama-3.2-1B      &  0.46 & 0.75 &  3.59 & 0.82 &  0.62 & 0.81 &  0.28 & 0.46 & 27.99 &  4.49 \\
\multicolumn{1}{r}{+RAG}  & 16.92 & 6.24 & 15.71 & 4.82 & 15.61 & 5.98 & 11.74 & 2.58 & 18.29 &  4.13 \\
\addlinespace[2pt]
Qwen-2.5-1.5B      &  0.24 & 2.03 &  2.81 & 4.64 &  0.22 & 2.18 &  0.31 & 1.23 & 30.89 & 33.86 \\
\multicolumn{1}{r}{+RAG}  & 21.92 &15.57 & 27.12 &15.71 & 22.03 &15.24 & 18.60 &11.74 & 19.49 & 10.17 \\
\addlinespace[2pt]
Qwen-2.5-7B        &  0.86 & 3.04 &  4.66 & 6.22 &  0.78 & 2.93 &  0.84 & 1.30 & 44.28 & 40.10 \\
\multicolumn{1}{r}{+RAG}    & 25.77 &20.09 & 31.97 &24.00 & 25.68 &19.49 & 25.92 &16.03 & 35.42 & 23.02 \\
\addlinespace[2pt]
Gemma-3-1B-pt     &  0.51 & 0.60 &  4.33 & 0.65 &  0.58 & 0.64 &  0.31 & 0.62 & 26.58 &  1.92 \\
\multicolumn{1}{r}{+RAG} & 14.43 & 5.59 & 16.59 & 5.04 & 13.64 & 5.66 &  8.13 & 2.16 & 19.96 &  5.03 \\
\bottomrule
\end{tabular}
\end{adjustbox}
\vspace{-5mm}
\end{table}

%% file: Sections/4_Train.tex
\section{Knowledge Injection Approaches}\label{sec:approach}

In this paper, we focus on comparing approaches that can internalize new knowledge into model parameters with separate discussion on RAG in Table~\ref{tab:merged_static_rag}. Specifically, we continue to pre-train various model architectures with different training objectives as introduced in Section~\ref{sec:objective_clm} and Section~\ref{sec:objective_bilm}. Finally, in Section~\ref{sec:ensemble_pipeline}, we introduce a modular approach that treats BiLMs as knowledge repositories and integrates them with LLMs.

\subsection{Knowledge Injection Preliminaries}

Knowledge injection~\citep{fu2023revisiting,zhang2023plug,ovadia2023fine,onoe2023can,xu2025memorizing} assumes that we have a pre-trained model ${\cal M}$ and a set of knowledge ${\cal T}=\{t\}$. The model is then trained on the knowledge via a training objective ${\cal M'}={\cal L}({\cal T}; {\cal M})$ to integrate ${\cal T}$ into ${\cal M}$. To conduct standardized comparisons, we define a \emph{knowledge upsampling} parameter $s\in \mathbb{N}^+$ as the number of times a single knowledge entry $t\in {\cal T}$ is encountered during training, thereby controlling the amount of training.

\subsection{Continued Pretraining for CLMs}\label{sec:objective_clm}

\textbf{Next Token Prediction} We continue to pre-train the LLM on raw textual knowledge with next-token-prediction objective that maximizes the log-likelihood of $\sum_{i=1}^l\log p(t_{i+1}|t_1,\cdots,t_{i})$ for a text sequence $t$ with token length $l$. We upsample by replicating each $t$ for $s$ times during training.

\textbf{Synthetic QA Training} Inspired by the approach proposed in~\cite{wang2025selfupdatable}, we prompt \texttt{gpt-4.1-mini} to convert the factual knowledge into all possible forms of questions (see prompts in Appendix~\ref{sec:prompts}). We then fine-tune the LLMs to predict the answer conditioned on the questions. Notice that here we use an external model for QA generation for its simplicity and quality. It is also able to use the corresponding instruct versions of open-source models as described in~\cite{wang2025selfupdatable}. We upsample from the generated set of training QAs through replication to form the final training set.

\textbf{Span Prediction} In order to align with the training objective of BiLMs, we propose span prediction tasks for CLMs. Specifically, we format each input with a mask prediction prompt: ``\textit{Predict the masked words in the following sentence: \{input\_str\}\textbackslash nMasked words:\textbackslash n}'' where the input string is a corrupted text and the target is the span that recovers it. For a fair comparison with BiLMs, we employ the same masking strategy and upsampling as introduced in Section~\ref{sec:objective_bilm}. At test time, we use the same prompt template and append a mask token after the question.

\subsection{Continued Pretraining for BiLMs}\label{sec:objective_bilm}

\textbf{Span Prediction} We employ the span prediction objective from T5~\citep{raffel2020exploring}. Specifically, it maximizes the following log-likelihood during training $\sum_{(i,s)\in {\cal S}}\log p(t_i,\cdots,t_{i+s}|t_1,\cdots,t_{i-1},t_{i+s+1},\cdots,t_{l})$ for a random masking strategy ${\cal S}$. The upsampling parameter is then $s=|\cal{S}|$. At test time, we simply append an extra token after the question in order to predict the answer to the question.

\textbf{Exhaustive Masking Strategy} In order to improve the sampling efficiency, we design a simple exhaustive sampling strategy. Specifically, we first generate all possible candidates of masked inputs given the minimum and maximum span lengths. During training, we generate a upsampled list of masked inputs based on the candidates. In this way, we guarantee the diversity of training samples while minimizing the upsampling parameter.

\subsection{Ensemble Pipeline}\label{sec:ensemble_pipeline}
Injecting an unbounded amount of new knowledge will inevitably diminish the effectiveness and incur catastrophic forgetting. Building on insights from modular architectures~\cite{mitchell2022memory,li2022branch,feng2023knowledge}, we propose a collaborative framework that coordinates multiple BiLMs as external knowledge repositories for LLMs. Our framework organizes external knowledge through two complementary partitioning strategies: (1) semantic clustering, where a Gaussian Mixture Model (GMM) groups facts into clusters based on their dense semantic embeddings, and (2) temporal clustering, which leverages fact timestamps to partition knowledge chronologically. To ensure robust routing, we train a scope classifier to discriminate between in-scope clusters (inter-class separation) and out-of-scope queries.
The classifier is optimized using binary cross-entropy loss, where we assign a uniform label of $0$ to all out-of-scope training instances. Negative training examples are derived from facts dated between 2004 and 2009. Each cluster is then internalized by a dedicated BiLM, forming a modular knowledge base.
During inference, queries are either routed to the most relevant BiLM or deferred to the base LLM if deemed out-of-scope by a confidence threshold. This design ensures that the LLM’s original knowledge remains intact, while injected knowledge is adaptively utilized through the BiLM ensemble, effectively mitigating catastrophic forgetting. Our framework thus enables synergistic integration of pretrained and external knowledge without compromising the LLM’s foundational capabilities.






%% file: Sections/5_Exp.tex
\begin{table}[t]
\centering
\caption{Main results of knowledge injection with full dataset. Best results are bolded, and second-best are underscored. \texttt{NTP} stands for next-token-prediction, \texttt{QA} for synthetic QA and \texttt{SP} for span prediction. We set $s=1,000$ for all experiments. More results of BiLMs can be found in Appendix.}
\label{tab:main_results}
\begin{adjustbox}{max width=0.9\textwidth}
\begin{tabular}{ll*{5}{cc}}
\toprule
\multirow{2}{*}{\textbf{Model}} &
\multirow{2}{*}{\textbf{Obj.}} &
\multicolumn{2}{c}{\textbf{Reliability}} &
\multicolumn{2}{c}{\textbf{Generality}} &
\multicolumn{2}{c}{\textbf{Paraphrase}} &
\multicolumn{2}{c}{\textbf{Portability}} &
\multicolumn{2}{c}{\textbf{Locality}} \\ 
\cmidrule(lr){3-4}\cmidrule(lr){5-6}\cmidrule(lr){7-8}\cmidrule(lr){9-10}\cmidrule(l){11-12}
 & & Match & F1 & Match & F1 & Match & F1 & Match & F1 & Match & F1 \\
\midrule
\multirow{3}{*}{Llama‑2‑7b} 
  & NTP          & 0.90 & 1.04 & 7.20 & 1.42 & 1.09 & 0.99 & 0.58 & 0.66 & 25.74 & 2.67 \\ 
  & QA          & 9.02 & 13.56 & 18.38 & 28.58 & 4.86 & 9.33 & 0.98 & 4.22 & 44.60 & \textbf{50.17} \\ 
  & SP          & 10.93 & 13.90 & 13.45 & 14.61 & 8.40 & 11.46 & 1.60 & 3.55 & 42.15 & 41.41 \\ 
\midrule
\multirow{3}{*}{Llama‑3.1‑8B}
  & NTP          & 1.49 & 1.73 & 9.24 & 2.52 & 1.58 & 1.70 & 0.78 & 1.05 & \underline{45.32} & 8.72 \\ 
  & QA          & 6.90 & 11.21 & 15.70 & 25.77 & 3.72 & 8.12 & 1.02 & 4.06 & 39.44 & \underline{49.47} \\ 
  & SP          & 16.09 & 18.02 & 16.67 & 19.86 & 12.23 & 14.29 & 3.20 & 4.52 & 42.04 & 43.22 \\ 
\midrule
\multirow{3}{*}{Llama‑3.2‑1B}
  & NTP          & 0.45 & 1.65 & 0.54 & 7.75 & 0.43 & 1.29 & 0.36 & 0.64 & 0.51 & 6.88 \\ 
  & QA          & 16.92 & 18.84 & 29.80 & 41.35 & 13.83 & 15.95 & 1.55 & 2.59 & 4.81 & 5.66 \\ 
  & SP          & 3.03 & 5.88 & 4.96 & 5.95 & 1.72 & 4.35 & 0.55 & 2.40 & 16.46 & 22.34 \\ 
\midrule
\multirow{3}{*}{Qwen‑2.5‑1.5B}
  & NTP          & 0.74 & 3.30 & 1.00 & 12.40 & 0.69 & 2.92 & 0.50 & 0.98 & 1.28 & 19.06 \\ 
  & QA          & 19.08 & 4.41 & \textbf{36.16} & 7.36 & 15.83 & 3.85 & \underline{4.62} & 0.97 & 12.16 & 4.72 \\ 
  & SP          & 1.06 & 1.03 & 7.21 & 1.43 & 1.13 & 1.03 & 0.39 & 1.09 & 21.80 & 2.38 \\ 
\midrule
\multirow{3}{*}{Qwen‑2.5‑7B}
  & NTP          & 0.46 & 1.21 & 5.04 & 2.19 & 0.58 & 1.17 & 0.64 & 1.59 & 28.71 & 32.51 \\ 
  & QA          & 2.29 & 3.91 & 8.61 & 9.24 & 0.74 & 3.23 & 0.39 & 2.48 & 29.08 & 29.92 \\ 
  & SP          & 2.01 & 1.72 & 2.73 & 9.00 & 1.83 & 1.46 & 1.86 & 0.64 & 10.20 & 30.31 \\ 
\midrule
\multirow{3}{*}{Gemma‑3‑1B}
  & NTP          & 0.44 & 1.14 & 0.60 & 5.93 & 0.47 & 1.03 & 0.32 & 0.60 & 0.39 & 5.94 \\ 
  & QA          & 8.17 & 11.81 & 15.28 & 23.96 & 4.06 & 7.74 & 1.04 & 3.98 & 27.19 & 31.73 \\ 
  & SP          & 7.37 & 10.41 & 9.61 & 11.50 & 5.50 & 8.32 & 1.62 & 3.98 & 28.04 & 32.18 \\ 
\midrule
Flan-T5-220M  & \multirow{4}{*}{SP}
                & 10.06 & 13.38 & 7.05 & 10.22 & 6.10 & 9.47 & 0.16 & 0.65 & 4.55 & 6.84 \\
Flan-T5-220M (ens) &
                & 39.16 & 41.96 & 20.96 & 23.96 & 25.72 & 29.10 & 2.26 & 3.66 & 44.59 & 9.40 \\
Flan-T5-770M &
                & \underline{46.09} & \underline{48.83} & 25.58 & \underline{29.60} & \underline{33.25} & \underline{36.70} & 3.86 & \underline{6.70} & 15.47 & 16.22 \\
Flan-T5-770M (ens) &
                & \textbf{52.82} & \textbf{53.85} & \underline{31.84} & \textbf{34.91} & \textbf{40.02} & \textbf{42.04} & \textbf{6.56} & \textbf{8.14} & \textbf{49.58} & 11.55 \\
\bottomrule
\end{tabular}
\end{adjustbox}
\vspace{-5mm}
\end{table}

\section{Results and Analysis}

\subsection{Experimental Setup}\label{sec:exp_setup}

For \textbf{CLMs}, we train models from $3$ model families including \texttt{Llama-2/3}~\citep{touvron2023llama,grattafiori2024llama}, \texttt{Qwen2.5}~\citep{yang2024qwen2} and \texttt{Gemma3}~\citep{team2025gemma} with different sizes using the objectives described in Section~\ref{sec:approach}.
We choose base models for knowledge injection since these mod
For comparison, we use full-parameter training for models less than $3$B and use LoRA~\citep{hu2022lora} with rank $32$ and $\alpha=16$ for other models.  For learning rate, we use $2e-5$ for full-parameter training and $2e-4$ for LoRA training.
For \textbf{BiLMs}, we choose \texttt{Flan-T5-220M/770M}~\citep{chung2024scaling}, T5(v1.1)-large~\citep{raffel2020exploring} and Roberta-large~\cite{liu2019roberta}. We train the full parameters for all BiLM models. For \texttt{Flan-T5-220M}, we choose a learning rate of $3e-4$ and for large versions of T5 models we use $1e-4$.
Other hyperparameters can be found in Appendix~\ref{sec:hyper}.
Each trained model is then evaluated with all types of questions in the same way as mentioned in Section~\ref{sec:evaluation}. Notice that for \texttt{Roberta-large}, we append fixed $10$ mask tokens after the question for generation.
For ensemble models, we train \texttt{DeBERTa-v3-large}~\citep{he2021debertav3} as our scope classifier and we integrate the injected BiLMs with \texttt{Llama-3.1-8B}. 

\subsection{Main Results}
We demonstrate five insights below extracted from the main results in Table~\ref{tab:main_results}.

\noindent \textbf{NTP objective is not suitable for knowledge injection.}
Notably, results on first four types of questions after NTP training are mostly lower than $1\%$ for match accuracy or even token F1. These results are even lower than the static analysis in Table~\ref{tab:merged_static_rag}.
We also observe catastrophic forgetting according to the locality performance after training. For example the locality match is decreased by $25.62\%$ for \texttt{Llama-2-7b}.
The poor generalizability can be attributed to both the formatting difference between training and evaluation, and the low context visibility for causal attention mask.

\noindent \textbf{BiLMs are much more effective.}
The performances of both \texttt{Flan-T5-220M/770M} with span prediction are presented as a showcase for BiLMs. Despite smaller scales ($220$M for base version and $770$M for large version), we found these models to be much more effective than CLMs. For instance, \texttt{Flan-T5-770M} achieves $46.09\%$ match accuracy on reliability, which reflects that the model can memorize almost half of the knowledge correctly. However, it is still not clear whether the effectiveness is derived from the diverse training examples produced by the random masking strategy or the architectural advantage. Thus it is important to compare the performance under controlled experiment on training objective.

\noindent \textbf{Effectiveness of BiLMs may come from architecture rather than training objective.}
Results from synthetic QA and span prediction shows notable improvements against NTP baseline, and the former is usually more effective than the latter. For example, the reliability match is improved from $0.45$ to $16.92$ for \texttt{Llama-3.2-1B}. With span prediction, paraphrase match performance is improved by $7.31\%$ for \texttt{Llama-2-7b}. More importantly, comparing span prediction results on both CLMs and \texttt{Flan-T5} models, we see the significant superiority on the latter for the first four types of questions, especially considering the aligned training objective and number of upsampling during training. This directly shows that the performance gain of BiLMs may be related with architectural advantage and we encourage further analysis on this matter (see discussion in Section~\ref{sec:discuss}).

\noindent \textbf{Ensemble pipeline can further improve the performance.}
We show the ensemble results with $10$ \texttt{Flan-T5} models and the rejected questions will be answered by \texttt{Llama-3.1-8B}. As shown in Table~\ref{tab:main_results}, ensemble models further improve the performance by $29.1\%$ match accuracy on reliability for \texttt{Flan-T5-220M} and $6.73\%$ for the large version. Furthermore, the match accuracy of locality is significantly improved for both versions of \texttt{Flan-T5}.
We attribute the performance gain to the dedicated training on the assigned clusters and the cooperation between LLMs and the trained BiLMs. We show further analysis in Section~\ref{sec:ensemble_exp}.

\noindent \textbf{Knowledge association shows less improvements.} As can be observed in the table, the results for portability is less improved compared with other types of questions for all models and training objectives. Similar phenomenon is also observed in~\cite{xu2025memorizing}, who found that continued pre‑training reliably recalls edited triples but fails on derivative association queries, and by~\cite{zhong2023mquake}, where accuracy drops from nearly 90\% on single‑hop recall to below 15\% on two‑hop association.


\subsection{Effect of Amount of Knowledge Injected}
\begin{figure}[t]
    \centering
    \includegraphics[width=\linewidth]{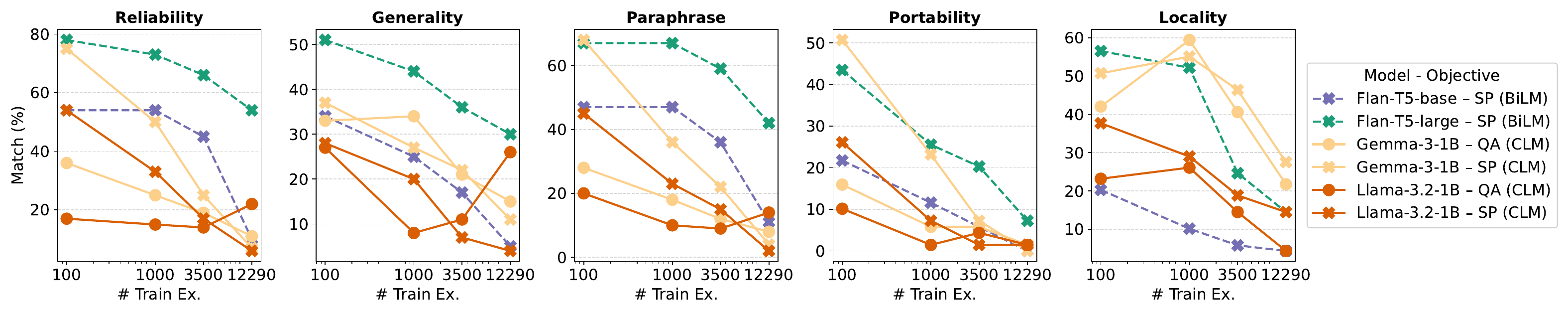}
    \caption{Effect of number of knowledge injected with number of upsampling $s=1000$.}
    \label{fig:number_of_data}
    \vspace{-4mm}
\end{figure}

In order to understand the capacity of each model under different training objectives, we analyze the amount of knowledge injected in Figure~\ref{fig:number_of_data}. Specifically, we train the models with the first $100/1,000/3,500$ knowledge entries or the full dataset. For fair comparison, we evaluate all the models with the questions generated from the first $100$ knowledge entries. As the number of training data progressively increases from $100$ to $3500$, the reliability and paraphrase for both \texttt{Gemma-3-1B} and \texttt{Llama-3.2-1B} decreases sharply while the performances for \texttt{Flan-T5} models stay relatively constant. However, the performance decreased significantly after injecting the full dataset especially for \texttt{Flan-T5-220M} which demonstrates the capacity limit of it. While single model capacity is always constrained, our ensemble pipeline can alleviate the issue by combining predictions from multiple specialized models. This approach not only compensates for the capacity limitations of smaller models like \texttt{Flan-T5-220M} but also leverages complementary strengths across architectures.

\begin{figure}[t]
    \centering
    \includegraphics[width=\linewidth]{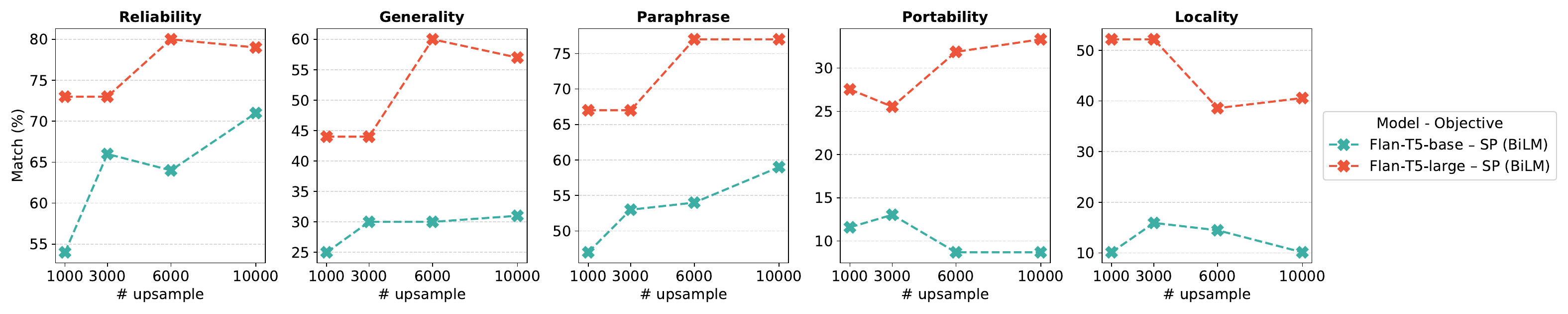}
    \caption{Effect of number of upsampling with $1000$ injected knowledge.}
    \label{fig:upsampling}
    \vspace{-5mm}
\end{figure}

\subsection{Effect of Number of Upsampling}
To understand whether further increasing the number of upsampling could help the performance, we further conduct controlled experiments by training with first $1000$ knowledge entries and increased number of upsampling. We found the performance for the first four question types further increases with increased number of upsampling except for the portability performance for \texttt{Flan-T5-220M} which stays between $5-15\%$. For instance, the match accuracy of paraphrase further increases by up to $15\%$ with increased upsampling. However, the performance saturates after $6000$ upsampling for most question types. We can also observe that locality performance is decreased by up to $10\%$ after $3000$ upsampling. For efficiency purpose, we control the number of upsampling in the main experiment as $1000$.

\subsection{Ablation Study on Clustering and Scope Classifier}\label{sec:ensemble_exp}
To better understand the effect of clustering and scope classifier quality, we conduct ablation study in Table~\ref{tab:ablation_ensemble}. Specifically, we ablate the performance between two kinds of the clustering algorithms: temporal and semantic. We found that temporal cluster consistently performs worse than semantic one. Furthermore, the performance for semantic clustering is improving with more number of clusters while the same is reversed for temporal one. We assume that this might be attributed to the failures of two possible components: scope classifier and knowledge injection. By using ground truth classifier in the ensemble pipeline, we ablate the effect of scope classifier, which can be found with ``-perfect'' results. The ``Temporal-perfect'' performs similarly with ``semantic-perfect'' which demonstrates that it is mainly the classifier that affects the performance of temporal clustering.
\begin{table}[t]
\centering
\caption{Ablation study on ensemble models using \texttt{Flan-T5-220M} as the base model.}
\label{tab:ablation_ensemble}
\begin{adjustbox}{max width=0.9\textwidth}
\begin{tabular}{lc*{5}{cc}}
\toprule
\multirow{2}{*}{\textbf{Type}} &
\multirow{2}{*}{\textbf{\# clusters}} &
\multicolumn{2}{c}{\textbf{Reliability}} &
\multicolumn{2}{c}{\textbf{Generality}} &
\multicolumn{2}{c}{\textbf{Paraphrase}} &
\multicolumn{2}{c}{\textbf{Portability}} &
\multicolumn{2}{c}{\textbf{Locality}} \\ 
\cmidrule(lr){3-4}\cmidrule(lr){5-6}\cmidrule(lr){7-8}\cmidrule(lr){9-10}\cmidrule(l){11-12}
 & & Match & F1 & Match & F1 & Match & F1 & Match & F1 & Match & F1 \\
\midrule
\multirow{3}{*}{Temporal} 
  & 3          & 17.76 & 20.52 & 11.14 & 13.85 & 9.98 & 12.64 & 0.82 & 1.51 & 40.47 & 7.32 \\ 
  & 5          & 15.18 & 17.41 & 11.49 & 14.00 & 8.79 & 11.12 & 0.84 & 1.46 & 35.48 & 7.21 \\ 
  & 10          & 9.06 & 10.02 & 9.73 & 8.73 & 5.51 & 6.40 & 0.94 & 1.13 & 40.47 & 1.13 \\ 
\midrule
\multirow{3}{*}{Temporal-perfect} 
  & 3          & 27.13 & 30.42 & 15.23 & 18.57 & 18.23 & 21.74 & 1.32 & 3.21 & 49.58 & 7.66 \\ 
  & 5          & 38.59 & 41.95 & 22.06 & 25.73 & 27.00 & 30.79 & 3.08 & 5.86 & 49.58 & 7.66 \\ 
  & 10          & 44.24 & 47.23 & 27.12 & 30.19 & 32.87 & 36.57 & 6.46 & 9.82 & 49.58 & 7.66 \\ 
\midrule
\multirow{3}{*}{Semantic}
  & 3          & 28.36 & 31.55 & 14.36 & 17.62 & 17.28 & 20.42 & 1.12 & 1.83 & 44.15 & 2.43 \\ 
  & 5          & 32.16 & 34.86 & 16.94 & 19.79 & 20.63 & 23.46 & 1.55 & 2.48 & 45.02 & 5.00 \\ 
  & 10          & 39.16 & 41.96 & 20.96 & 23.96 & 25.72 & 29.10 & 2.26 & 3.66 & 44.59 & 9.40 \\ 
\midrule
\multirow{3}{*}{Semantic-perfect}
  & 3          & 29.70 & 33.25 & 15.54 & 19.18 & 19.06 & 22.80 & 1.29 & 3.03 & 49.58 & 7.66 \\ 
  & 5          & 35.88 & 39.03 & 19.11 & 22.62 & 24.08 & 27.59 & 2.79 & 5.64 & 49.58 & 7.66 \\ 
  & 10          & 44.23 & 47.23 & 27.12 & 30.19 & 32.87 & 36.57 & 6.46 & 9.82 & 49.58 & 7.66 \\ 
\bottomrule
\end{tabular}
\end{adjustbox}
\vspace{-3mm}
\end{table}

\subsection{Case Study}
We show the case study for one of our trained model using \texttt{Flan-T5-770M} in Table~\ref{tab:case_study}. Despite low lexical similarity between the original fact and question, we still see that the predicted answers can match the ground truths, especially for the third example which shows a portability question. However, we do observe wrong matches, for example the last example in the table shows a hallucinated prediction against the ground truth.
\definecolor{darkgreen}{RGB}{0,100,0}   
\definecolor{darkred}  {RGB}{139,0,0}   
\begin{table}[t]
\centering
\footnotesize
\setlength{\tabcolsep}{4pt}
\renewcommand{\arraystretch}{0.9}
\caption{Case study on the model predictions. The backbone model is \texttt{Flan-T5-770M}, trained on the full dataset with $s = 3{,}000$. Correct predictions are highlighted in green, and incorrect ones in red.}
\label{tab:case_study}
\begin{adjustbox}{max width=0.9\linewidth}
\begin{tabularx}{\linewidth}{@{}XXll@{}}
\toprule
\textbf{Fact} & \textbf{Question} & \textbf{Ground Truth} & \textbf{Prediction}\\
Mary Healy, an international speaker on faith healing in the Catholic Church, only became interested in the subject during her 2014 sabbatical & Who is an international speaker on faith healing in the Catholic Church who became interested in the subject during a 2014 sabbatical? & Mary Healy & \textcolor{darkgreen}{Mary Healy} \\
\midrule
AI expert Tess Posner resigned her role as a CEO in order to concentrate on her music career        & Who stepped down from being a CEO to pursue their music career? & Tess Posner & \textcolor{darkgreen}{AI expert Tess Posner} \\
\midrule
Journalist Jack Berry was influential in lifting the ban on female reporters in the locker room at The Masters          & I've been reading about a storied men's golf championship held every spring in Augusta, Georgia, ... Who was that journalist? & Jack Berry & \textcolor{darkgreen}{Jack Berry} \\
\midrule
Washington State Route 304 was accidentally removed ... for two years & How long was Washington State Route 304 accidentally removed from the state highway system? & two years & \textcolor{darkred}{three years}\\
\bottomrule
\end{tabularx}
\end{adjustbox}
\vspace{-5mm}
\end{table}

%% file: Sections/6_Discuss.tex
\section{Conclusion and Discussion}\label{sec:discuss}
In this work, we introduced {\dataset}, a novel real-world, large-scale benchmark for knowledge injection that autonomously evolves over time, eliminating the need for manual updates. To rigorously evaluate knowledge capabilities in language models, we designed a multi-dimensional evaluation suite structured as question-answering tasks, probing both knowledge memorization and associative reasoning.
Our extensive experiments reveal a critical limitation: under continued pre-training, Causal Language Models (CLMs) exhibit significantly weaker knowledge memorization compared to Bidirectional Language Models (BiLMs). To address this gap, we proposed a modular collaborative framework that integrates BiLMs as dynamic external knowledge repositories with LLMs. This approach not only compensates for CLMs’ limitations but also achieves a $29.1\%$ improvement in reliability, demonstrating the viability of model ensembles for knowledge-intensive tasks.


\noindent \textbf{BiLMs vs. CLMs}
CLMs become the prevailing choice for LLM architecture. This is largely due to the low-latency generation and simple architecture design. In our experiments, we show that bidirectional attention, when paired with fill-in-the-blank type objective enable models to capture richer dependencies between tokens by leveraging both past and future contexts. Our findings suggest that bidirectional architectures excel in scenarios requiring dense knowledge integration, such as entity disambiguation, factual reasoning, or structured data understanding. However, these results do not negate the advantages of CLMs in generation tasks but instead highlight opportunities for hybrid architectures. uture work might explore dynamic attention mechanisms that adaptively toggle between bidirectional and unidirectional modes, or modular designs where specialized bidirectional components handle knowledge-intensive subtasks as suggested in the paper.

%% file: Sections/7_Appendix.tex
\section{More Details in Experiments}\label{sec:hyper}
We illustrate the details of hyperparameters selected in this paper. For all the experiments, we use up to NVIDIA-A100 80G GPU for training and evaluation. We train for $1$ epoch if not specified. We use a minimum span length of 1 and maximum of $5$.

\noindent \textbf{Next-token-prediction}
We train with NTP objective on $8$ GPUs where the batch size is $256$ and upsampling is $1000$. The learning rate and model configurations are described as in Section~\ref{sec:exp_setup}.

\noindent \textbf{Synthetic QA}
We train with synthetic QA objective on $8$ GPUs where the batch size is $256$ and upsampling is $1000$. The learning rate and model configurations are described as in Section~\ref{sec:exp_setup}.

\noindent \textbf{Span Prediction}
For CLMs, we train with span prediction objective on $4$ GPUs where the batch size is $128$ and upsampling is $1000$. The learning rate and model configurations are described as in Section~\ref{sec:exp_setup}.
For BiLMs, we train with span prediction on $2/4$ GPUs where the batch size $128$ and upsampling is $1000$. The learning rate and model configurations are described as in Section~\ref{sec:exp_setup}.

\noindent \textbf{Scope Classifier}
For scope classifier, we train with \texttt{DeBERTa-large-v3} where we set the learning rate $2e-5$ and batch size $128$. We train for $10$ epochs.

\noindent \textbf{Metrics}
We evaluate model performance using two metrics: (1) Match, a binary indicator that returns true if the target string is a substring of the model’s output (case-sensitive), and false otherwise; and (2) Token-Level F1, the harmonic mean of precision and recall computed by aligning the model’s output tokens with the target tokens. The Match metric assesses strict presence of the target sequence, while the Token F1 score quantifies partial lexical overlap, offering complementary insights into generation quality. Both metrics are computed using whitespace-based tokenization to ensure consistency with standard text generation benchmarks.

\section{Prompts for Question Generation}\label{sec:prompts}

\subsection{Reliability Question Generation Prompt}

We use \texttt{GPT-4o} for generating reliability QAs with the following prompt:

\begin{lstlisting}[language=TeX,caption={Prompt for reliability QA Generation}]
Given a DYK fact in JSON format containing 'text' and 'bold_entity' fields, generate a question. Your output should be a JSON object containing the question with its corresponding answer. Your response should follow these criteria:

1. The question should be answerable using only the information provided in the fact
2. The answer should be the bold_entity
3. The question should be clear, natural, and specific so that the answer can be easily identified (i.e., use as many details as possible from the fact)
4. The bold entity should not be mentioned in the question since it is the answer. But make sure that the question's answer is the bold entity.

Example:
Input:
{{
   'text': 'that Margrit Waltz has ferried planes to points on five continents?',
   'bold_entity': 'Margrit Waltz',
}}

Expected output:
{{
    "question": {{
        "text": "Who has ferried planes to points on five continents?",
        "answer": "Margrit Waltz"
    }}
}}

Now please generate a question with answer for this fact:
{test_example}
\end{lstlisting}

\subsection{Paraphrase Question Generation Prompt}
We use \texttt{GPT-4.1} for generating paraphrase questions based on the reliability question with the following prompt:
\begin{lstlisting}[language=TeX,caption={Prompt for paraphrase QA generation}]
Given a pair of question and answer, generate three different paraphrases of the question. Make sure the answer is the same as before. Your output should be a JSON object with a list of dictionaries under the key "paraphrases". Each dictionary should have a "question" key and an "answer" key. Here is the pair of question and answer:

Question: {question}
Answer: {answer}
\end{lstlisting}
We pick the first paraphrased question from the generated list.

\subsection{Generality Question Generation Prompt}
We use \texttt{GPT-4.1} for generating generality questions based on the reliability question and the fact with the following prompt:
\begin{lstlisting}[language=TeX,caption={Prompt for generality QA generation}]
Given a pair of question and answer, generate three different alternative questions. Make sure the question asks about a different aspect of the same fact. Remember to follow the rules below:

1. The answer is one aspect of the fact (such as an entity / year / number etc.) apart from the original answer.
2. The answer should be concise and direct without any redundant words. And it shoud be a part of the fact.
3. The question should utilize all the information in the fact and be specific.
4. Do not use any information that is beyond the fact.
5. Your output should be a JSON object with a list of dictionaries under the key "alternatives". Each sub-dictionary should have a "question" key and an "answer" key.

Here is the pair of question and answer:

Fact: {fact}
Question: {question}
Answer: {answer}
\end{lstlisting}

\subsection{Portability Question Generation Prompt}
To generate the portability questions, we need to first identify an entity from the original knowledge entry that is non-bolded and have an associated Wikipedia article. After that, we prompt the \texttt{o3-mini} to first generate an entity description given its Wikipedia link (Notice that we discard those knowledge entries where the entity can not be found or the link is broken). The following is the entity description prompt:
\begin{lstlisting}[language=TeX,caption={Prompt for entity description generation}]
Replace the entity name with a description of it without mentioning the entity name. The description should be unique and specific. Make sure that you can infer the entity name using the description. You might also be provided with the wikipedia page of the entity. The output should be a JSON object with the following format:

{{
    "description": "The description of the entity",
}}

Wikipedia page: {page}
Entity name: {entity}
\end{lstlisting}
Then, we further prompt \texttt{o3-mini} to generate multi-hope QA pairs by replacing the entity with the description based on the reliability question.
\begin{lstlisting}[language=TeX,caption={Prompt for portability QA generation}]
Below are a few examples of natural, scenario-based questions where a user describes a scenario and then asks a question:

Example 1:
Alternative description: "a historic European city known for its iconic architecture and cobblestone streets."
User's natural question: "I recently visited a charming European city famous for its unique architecture and quaint streets. Can you tell me about a famous monument there?"
Entity name: Paris

Example 2:
Alternative description: "a groundbreaking technology company that revolutionized communication with its innovative products."
User's natural question: "I've been reading about a tech company that changed how we communicate through its innovative gadgets. What product are they best known for?"
Entity name: Apple

Now, given the alternative description and the original question below, generate a new, natural, scenario-based question. The new question should describe a scenario without mentioning the original entity name and then ask the question in a natural, conversational manner.

Alternative description: {description}
Entity name: {entity}
Original question: {question}

The output should be a JSON object with the following format:

{{
    "question": "The modified question"
}}
\end{lstlisting}

\subsection{Locality Question Generation Prompt}

We use \texttt{gpt-4.1} for locality question generation based on the previously generated entity description.

\begin{lstlisting}[language=TeX,caption={Prompt for portability QA generation}]
You'll generate a question-answer pair based on the description of an entity.

For each statement, you'll return a JSON object containing:
1. "question": The question that corresponds to the statement
2. "answer": The answer to the question

Example outputs:

1. Input: Jupiter is the largest planet in our solar system.
Output:
{{
  "question": "What is the largest planet in our solar system?",
  "answer": "Jupiter"
}}
2. Input: The capital of France is Paris.
Output:
{{
  "question": "What is the capital of France?",
  "answer": "Paris"
}}

Entity: {entity}
Description: {description}
\end{lstlisting}

\subsection{Training QA Generation Prompt}

We use \texttt{gpt-4.1-mini} for training QA generation for the approach described in Section~\ref{sec:approach}.

\begin{lstlisting}[language=TeX,caption={Prompt for training QA generation}]
Given a context, please generate related questions as comprehensively as possible with corresponding answers. The question has to be based on the context and the answer should be a short phrase.
This is an example:
Context: A small coastal town has a beach known for its colorful sea glass. The town hosts an annual festival celebrating this unique feature with art and conservation efforts.
Question: What attracts tourists to the small coastal town
annually? Answer: The unique sea glass beach.
Question: What is celebrated at the town's annual festival?
Answer: The natural phenomenon of sea glass.
Question: What type of activities are featured at the festival?

Format your output in a JSON object like the one below:
{{
    "questions": [
        {{
            "question": "What attracts tourists to the small coastal town annually?",
            "answer": "The unique sea glass beach."
        }},
        {{
            "question": "What is celebrated at the town's annual festival?",
            "answer": "The natural phenomenon of sea glass."
        }},
        {{
            "question": "What type of activities are featured at the festival?",
            "answer": "Art and conservation efforts."
        }}
    ]
}}
Context: {fact}
\end{lstlisting}

\section{More Results}\label{sec:more_results}

We present more results in this section. First, we show the numerical results that correspond to Figure~\ref{fig:number_of_data} in Table~\ref{tab:ds100}-\ref{tab:ds3500}. Notice that we also experimented with MemoryLLM~\citep{memoryllm} and M+~\citep{wang2025mextendingmemoryllmscalable}. Results show that despite their claimed long-context, they are not able to utilize the questions for answer our evaluation questions, which results in lower performance compared with knowledge injection. We can also observe the performance of other BiLMs in Table~\ref{tab:ds1000} where we see that \texttt{T5-v1.1-large} also performs better than CLMs while lower then \texttt{Flan-T5} models. Thus, we only include \texttt{Flan-T5} models in our main results.
Second, we also show the performance of reliability questions from 2004 to 2009 in Table~\ref{tab:static_results_0409}. As can be observed, the performance decreases progressively with newer questions.
Finally, we show the performance of ablation on \texttt{Flan-T5-770M} in Table~\ref{tab:ablation_ensemble_large}. We can draw similar conclusions as in Section~\ref{sec:ensemble_exp}.

\begin{table}[H]
\centering
\caption{Results of knowledge injection with first $100$ knowledge entries.}
\label{tab:ds100}
\begin{adjustbox}{max width=\textwidth}
\begin{tabular}{ll*{5}{cc}}
\toprule
\multirow{2}{*}{\textbf{Model}} &
\multirow{2}{*}{\textbf{Obj.}} &
\multicolumn{2}{c}{\textbf{Reliability}} &
\multicolumn{2}{c}{\textbf{Generality}} &
\multicolumn{2}{c}{\textbf{Paraphrase}} &
\multicolumn{2}{c}{\textbf{Portability}} &
\multicolumn{2}{c}{\textbf{Locality}} \\ 
\cmidrule(lr){3-4}\cmidrule(lr){5-6}\cmidrule(lr){7-8}\cmidrule(lr){9-10}\cmidrule(l){11-12}
 & & Match & F1 & Match & F1 & Match & F1 & Match & F1 & Match & F1 \\
\midrule
\multirow{2}{*}{Llama‑3.1‑1B}
  & QA          & 17.00 & 19.66 & 27.00 & 35.71 & 20.00 & 24.96 & 10.14 & 15.96 & 23.19 & 33.00 \\ 
  & SP          & 54.00 & 49.47 & 28.00 & 26.27 & 45.00 & 41.86 & 26.09 & 22.53 & 37.68 & 38.89 \\ 
\midrule
\multirow{2}{*}{Gemma‑3‑1B}
  & QA          & 36.00 & 37.38 & 33.00 & 39.88 & 28.00 & 31.59 & 15.94 & 18.55 & 42.03 & 49.37 \\ 
  & SP          & 75.00 & 63.36 & 37.00 & 34.15 & 68.00 & 57.98 & 50.72 & 48.87 & 50.72 & 62.42 \\ 
\midrule
Flan-T5-220M  & \multirow{2}{*}{SP}
                & 56.00 & 58.69 & 34.00 & 33.73 & 47.00 & 49.83 & 21.74 & 28.49 & 20.29 & 20.87 \\
Flan-T5-770M &
                & 78.00 & 79.27 & 51.00 & 49.67 & 67.00 & 66.87 & 43.48 & 45.50 & 56.52 & 73.72 \\
\midrule
MemoryLLM  & \multirow{2}{*}{---}
                & 13.00 & 0.75 & 13.00 & 0.58 & 10.00 & 0.75 & 4.35 & 0.53 & 5.80 & 0.47 \\
Mplus &
                & 4.00 & 0.39 & 8.00 & 0.29 & 5.00 & 0.41 & 4.35 & 0.40 & 14.49 & 0.40 \\
\bottomrule
\end{tabular}
\end{adjustbox}
\end{table}

\begin{table}[H]
\centering
\caption{Results of knowledge injection with first $1000$ knowledge entries.}
\label{tab:ds1000}
\begin{adjustbox}{max width=\textwidth}
\begin{tabular}{ll*{5}{cc}}
\toprule
\multirow{2}{*}{\textbf{Model}} &
\multirow{2}{*}{\textbf{Obj.}} &
\multicolumn{2}{c}{\textbf{Reliability}} &
\multicolumn{2}{c}{\textbf{Generality}} &
\multicolumn{2}{c}{\textbf{Paraphrase}} &
\multicolumn{2}{c}{\textbf{Portability}} &
\multicolumn{2}{c}{\textbf{Locality}} \\ 
\cmidrule(lr){3-4}\cmidrule(lr){5-6}\cmidrule(lr){7-8}\cmidrule(lr){9-10}\cmidrule(l){11-12}
 & & Match & F1 & Match & F1 & Match & F1 & Match & F1 & Match & F1 \\
\midrule
\multirow{2}{*}{Llama‑3.2‑1B}
  & QA          & 15.00 & 18.39 & 8.00 & 15.47 & 10.00 & 13.03 & 1.45 & 7.33 & 26.09 & 32.00 \\ 
  & SP          & 33.00 & 29.87 & 20.00 & 20.01 & 23.00 & 21.28 & 7.25 & 11.02 & 28.99 & 37.13 \\ 
\midrule
\multirow{2}{*}{Gemma‑3‑1B-pt}
  & QA          & 25.00 & 30.08 & 34.00 & 41.27 & 18.00 & 22.97 & 5.80 & 9.72 & 59.42 & 62.95 \\ 
  & SP          & 50.00 & 39.66 & 27.00 & 24.44 & 36.00 & 30.73 & 23.19 & 21.24 & 55.07 & 49.46 \\ 
\midrule
roberta-large & \multirow{5}{*}{SP}
                & 3.00 & 15.55 & 14.00 & 12.84 & 3.00 & 11.80 & 0.00 & 0.00 & 0.00 & 3.29 \\
t5-large &
                & 32.00 & 34.88 & 19.00 & 22.28 & 28.00 & 31.91 & 8.70 & 10.22 & 24.64 & 29.13 \\
t5-v1.1-large &
                & 52.00 & 53.78 & 34.00 & 32.66 & 42.00 & 41.23 & 21.74 & 24.54 & 46.38 & 42.75 \\
Flan-t5-220M  & 
                & 54.00 & 57.56 & 25.00 & 28.97 & 47.00 & 50.57 & 11.59 & 15.50 & 10.14 & 15.30 \\
Flan-t5-770M &
                & 73.00 & 73.97 & 44.00 & 43.22 & 67.00 & 68.21 & 27.54 & 35.76 & 52.17 & 58.21 \\
\bottomrule
\end{tabular}
\end{adjustbox}
\end{table}

\begin{table}[H]
\centering
\caption{Results of knowledge injection with first $3500$ knowledge entries.}
\label{tab:ds3500}
\begin{adjustbox}{max width=\textwidth}
\begin{tabular}{ll*{5}{cc}}
\toprule
\multirow{2}{*}{\textbf{Model}} &
\multirow{2}{*}{\textbf{Obj.}} &
\multicolumn{2}{c}{\textbf{Reliability}} &
\multicolumn{2}{c}{\textbf{Generality}} &
\multicolumn{2}{c}{\textbf{Paraphrase}} &
\multicolumn{2}{c}{\textbf{Portability}} &
\multicolumn{2}{c}{\textbf{Locality}} \\ 
\cmidrule(lr){3-4}\cmidrule(lr){5-6}\cmidrule(lr){7-8}\cmidrule(lr){9-10}\cmidrule(l){11-12}
 & & Match & F1 & Match & F1 & Match & F1 & Match & F1 & Match & F1 \\
\midrule
\multirow{2}{*}{Llama‑3.2‑1B}
  & QA          & 14.00 & 16.21 & 11.00 & 16.79 & 9.00 & 13.56 & 4.35 & 8.88 & 14.49 & 20.00 \\ 
  & SP          & 17.00 & 20.74 & 7.00 & 9.66 & 15.00 & 17.46 & 1.45 & 6.41 & 18.84 & 26.75 \\ 
\midrule
\multirow{2}{*}{Gemma‑3‑1B-pt}
  & QA          & 19.00 & 22.34 & 21.00 & 29.55 & 12.00 & 14.25 & 5.80 & 9.28 & 40.58 & 47.54 \\ 
  & SP          & 25.00 & 24.04 & 22.00 & 23.64 & 22.00 & 21.89 & 7.25 & 12.00 & 46.38 & 45.65 \\ 
\midrule
Flan-t5-220M  & \multirow{2}{*}{SP}
                & 45.00 & 49.14 & 17.00 & 21.30 & 36.00 & 41.09 & 5.80 & 9.72 & 5.80 & 10.87 \\
Flan-t5-770M &
                & 66.00 & 66.33 & 36.00 & 37.68 & 59.00 & 59.87 & 20.29 & 28.58 & 24.64 & 37.25 \\
\bottomrule
\end{tabular}
\end{adjustbox}
\end{table}

\begin{table}[h]
\centering
\caption{Results for older reliability questions from 2004 to 2009.}
\label{tab:static_results_0409}
\begin{adjustbox}{max width=\textwidth}
\begin{tabular}{l *{6}{cc}}
\toprule
\multirow{2}{*}{\textbf{Model}}
  & \multicolumn{2}{c}{\textbf{2004}}
  & \multicolumn{2}{c}{\textbf{2005}}
  & \multicolumn{2}{c}{\textbf{2006}}
  & \multicolumn{2}{c}{\textbf{2007}}
  & \multicolumn{2}{c}{\textbf{2008}} 
  & \multicolumn{2}{c}{\textbf{2009}}\\
\cmidrule(lr){2-3} \cmidrule(lr){4-5}
\cmidrule(lr){6-7} \cmidrule(lr){8-9} \cmidrule(lr){10-11} \cmidrule(lr){12-13}
  & Match & F1 & Match & F1 & Match & F1 & Match & F1 & Match & F1 & Match & F1 \\
\midrule
\addlinespace[2pt]
Llama-3.1-8B      & 12.03 & 2.03 & 10.42 & 2.39 & 6.53 & 2.04 & 5.12 & 1.63 & 4.34 & 1.75 & 3.79 & 1.64 \\
\addlinespace[2pt]
Llama-3.2-1B      & 2.22 & 1.24 & 0.69 & 1.12 & 0.43 & 1.02 & 0.84 & 1.03 & 0.58 & 1.04 & 0.62 & 0.99 \\
\addlinespace[2pt]
Qwen-2.5-1.5B      & 2.65 & 6.08 & 1.39 & 5.49 & 0.60 & 4.44 & 0.43 & 3.64 & 0.43 & 3.73 & 0.33 & 3.50 \\
\addlinespace[2pt]
Qwen-2.5-7B        & 6.66 & 10.36 & 6.05 & 10.36 & 2.66 & 7.08 & 2.39 & 6.53 & 1.95 & 6.13 & 1.48 & 5.32 \\
\addlinespace[2pt]
Gemma-3-1B-pt     & 2.39 & 0.72 & 1.46 & 0.72 & 1.03 & 0.72 & 0.67 & 0.75 & 0.58 & 0.78 & 0.59 & 0.73 \\
\bottomrule
\end{tabular}
\end{adjustbox}
\vspace{-3mm}
\end{table}

\begin{table}[h]
\centering
\caption{Ablation study on ensemble models using \texttt{Flan-T5-770M} as the base model.}
\label{tab:ablation_ensemble_large}
\begin{adjustbox}{max width=\textwidth}
\begin{tabular}{lc*{5}{cc}}
\toprule
\multirow{2}{*}{\textbf{Type}} &
\multirow{2}{*}{\textbf{\# clusters}} &
\multicolumn{2}{c}{\textbf{Reliability}} &
\multicolumn{2}{c}{\textbf{Generality}} &
\multicolumn{2}{c}{\textbf{Paraphrase}} &
\multicolumn{2}{c}{\textbf{Portability}} &
\multicolumn{2}{c}{\textbf{Locality}} \\ 
\cmidrule(lr){3-4}\cmidrule(lr){5-6}\cmidrule(lr){7-8}\cmidrule(lr){9-10}\cmidrule(l){11-12}
 & & Match & F1 & Match & F1 & Match & F1 & Match & F1 & Match & F1 \\
\midrule
\multirow{3}{*}{Temporal} 
  & 3          & 33.50 & 35.78 & 24.50 & 27.41 & 22.82 & 25.35 & 3.18 & 4.41 & 45.02 & 10.71 \\ 
  & 5          & 19.92 & 22.38 & 17.81 & 21.09 & 12.95 & 15.48 & 2.02 & 2.83 & 35.48 & 5.69 \\ 
  & 10          & 13.43 & 14.17 & 13.66 & 13.06 & 8.85 & 9.71 & 1.54 & 1.91 & 40.47 & 7.17 \\ 
\midrule
\multirow{3}{*}{Temporal-perfect} 
  & 3          & 51.57 & 53.88 & 34.50 & 38.12 & 41.51 & 44.24 & 10.11 & 13.90 & 49.58 & 7.66 \\ 
  & 5          & 51.60 & 54.14 & 34.58 & 38.79 & 41.22 & 44.52 & 12.15 & 15.90 & 49.58 & 7.66 \\ 
  & 10          & 65.70 & 66.72 & 42.69 & 46.15 & 55.13 & 57.18 & 20.24 & 23.76 & 49.58 & 7.66 \\ 
\midrule
\multirow{3}{*}{Semantic}
  & 3          & 44.24 & 46.55 & 27.85 & 31.37 & 32.00 & 34.68 & 4.44 & 6.19 & 44.60 & 11.24 \\ 
  & 5          & 44.03 & 45.93 & 29.48 & 32.66 & 32.46 & 34.75 & 4.27 & 5.73 & 45.02 & 5.00 \\ 
  & 10          & 52.82 & 53.85 & 31.84 & 34.91 & 40.02 & 42.04 & 6.56 & 8.14 & 49.58 & 11.55 \\ 
\midrule
\multirow{3}{*}{Semantic-perfect}
  & 3          & 46.62 & 49.17 & 30.25 & 34.21 & 35.79 & 38.84 & 6.60 & 9.97 & 49.58 & 7.66 \\ 
  & 5          & 49.28 & 51.50 & 33.28 & 37.36 & 37.88 & 40.83 & 9.42 & 13.11 & 49.58 & 7.66 \\ 
  & 10          & 56.57 & 57.62 & 36.29 & 39.77 & 45.71 & 47.80 & 11.88 & 15.02 & 49.58 & 7.66 \\ 
\bottomrule
\end{tabular}
\end{adjustbox}
\vspace{-3mm}
\end{table}

\section*{Limitation}\label{sec:limitation}
We discuss the limitations in this section. Our work claims that BiLMs perform much better than the CLMs. However, we prove this assumption empirically with no theoretical guarantees. Furthermore, limited by the computing resource, we are not able to completely pre-train a BiLM and a CLM under the same set of hyperparameter and data. Instead, we choose the popular pre-trained models for experiments without further controlling the experiments.